\documentclass[sigconf]{acmart}

\AtBeginDocument{%
  \providecommand\BibTeX{{%
    \normalfont B\kern-0.5em{\scshape i\kern-0.25em b}\kern-0.8em\TeX}}}

\copyrightyear{2022} 
\acmYear{2022} 
\setcopyright{acmcopyright}\acmConference[CIKM '22]{Proceedings of the 31st ACM International Conference on Information and Knowledge Management}{October 17--21, 2022}{Atlanta, GA, USA}
\acmBooktitle{Proceedings of the 31st ACM International Conference on Information and Knowledge Management (CIKM '22), October 17--21, 2022, Atlanta, GA, USA}
\acmPrice{15.00}
\acmDOI{10.1145/3511808.3557447}
\acmISBN{978-1-4503-9236-5/22/10}

\usepackage{amsmath}
\usepackage{amsthm}
\usepackage{stmaryrd}
\usepackage{mathtools}
\usepackage{stmaryrd}
\usepackage[lined,boxed,commentsnumbered,linesnumbered,ruled]{algorithm2e}
\usepackage{multirow}
\usepackage{booktabs}       
\newtheorem{thm}{Theorem}
\newtheorem{thm2}{Lemma}

\newtheorem{observation}[thm2]{Observation}
\newtheorem{definition}{Definition}

\usepackage{color}
\usepackage{xcolor}

\usepackage{enumitem}
\usepackage{balance}
\setlist[itemize]{leftmargin=*}
\usepackage{xspace}
\newcommand{\method}{\texttt{LE}\xspace }

\newcommand{\calG}{\mathcal{G}}
\newcommand{\calV}{\mathcal{V}}

\newcommand{\calE}{\mathcal{E}}
\newcommand{\calS}{\mathcal{S}}
\newcommand{\calT}{\mathcal{T}}
\newcommand{\bfH}{\mathbf{H}}
\newcommand{\bfD}{\mathbf{D}}
\newcommand{\bfA}{\mathbf{A}}
\newcommand{\bfP}{\mathbf{P}}
\newcommand{\bfI}{\mathbf{I}}
\newcommand{\bfX}{\mathbf{X}}
\newcommand{\bfY}{\mathbf{Y}}

\begin{document}

\title[Semi-supervised Hypergraph Node Classification on Hypergraph Line Expansion]{Semi-supervised Hypergraph Node Classification \\ on Hypergraph Line Expansion}

\author{Chaoqi Yang}
\affiliation{
  \institution{University of Illinois Urbana-Champaign}
  \country{}
  \institution{chaoqiy2@illinois.edu}
}

\author{Ruijie Wang}
\affiliation{
  \institution{University of Illinois Urbana-Champaign}
  \country{}
  \institution{ruijiew2@illinois.edu}
}

\author{Shuochao Yao}
\affiliation{
  \institution{George Mason University}
  \country{}
  \institution{shuochao@gmu.edu}
}

\author{Tarek Abdelzaher}
\affiliation{
  \institution{University of Illinois Urbana-Champaign}
  \country{}
  \institution{zaher@illinois.edu}
}


\renewcommand{\shortauthors}{Chaoqi Yang, Ruijie Wang, Shuochao Yao, \& Tarek Abdelzaher}

\begin{abstract}
	Previous hypergraph expansions are solely carried out on either vertex level or hyperedge level, thereby missing the symmetric nature of data co-occurrence, and resulting in  information loss. 
	To address the problem, this paper treats vertices 
	and hyperedges equally and proposes a new hypergraph expansion named the \emph{line expansion} (\method) for hypergraphs learning.
	 The new expansion bijectively induces a homogeneous structure from the hypergraph by modeling vertex-hyperedge pairs. Our proposal essentially reduces the hypergraph to a simple graph, which enables the existing graph learning algorithms to work seamlessly with the higher-order structure. We further prove that our line expansion is a unifying framework over various hypergraph expansions. We evaluate the proposed \method on five hypergraph datasets in terms of the hypergraph node classification task. The results show that our method could achieve at least $2$\% accuracy improvement over the best baseline consistently.
	 
	
\end{abstract}



\maketitle

\section{Introduction}
This paper proposes a new hypergraph structure transformation, namely \emph{line expansion} (\method), for the problem of hypergraph learning. Specifically, this paper focuses on hypergraph node classification.
The proposed \method is a topological mapping,  transforming the hypergraph into a homogeneous structure,  while preserving all the higher-order relations. \method allows all the existing graph
learning algorithms to work elegantly on hypergraphs.

The problem of hypergraph learning
is important. Graph structured data are ubiquitous in practical machine/deep learning applications, such as social networks \cite{ChitraR19}, protein networks \cite{klamt2009hypergraphs}, and co-author networks \cite{ZhouHS06}. Intuitive pairwise connections among nodes  are usually insufficient for capturing real-world higher-order relations. For example, in co-author networks, the edges between authors are created by whether they have co-authored a paper or not. A simple graph structure cannot separate the co-author groups for each paper. For another example, in biology, proteins are bound by poly-peptide chains, thus their relations are naturally higher-order. Hypergraphs allow modeling such multi-way relations, where (hyper)edges can connect to more than two nodes.

\begin{figure}[t]
	\centering
	\includegraphics[width=0.45\textwidth]{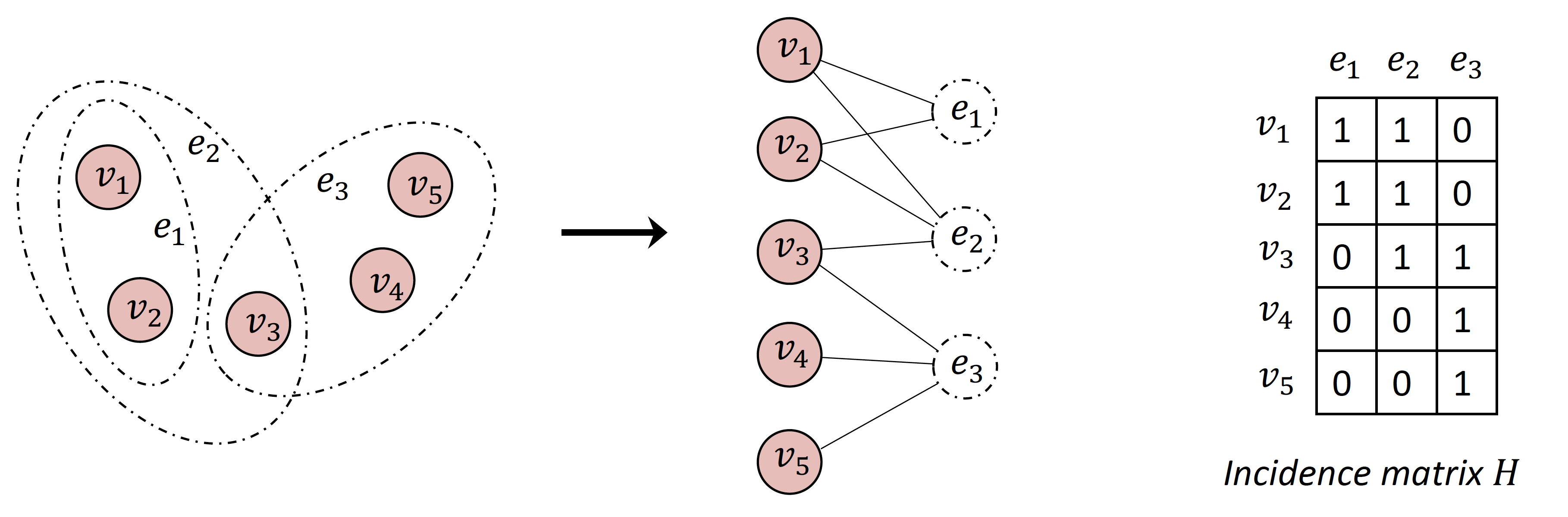}
	\vspace{-1mm}
	\caption{Bipartite Relation in Hypergraphs. {\normalfont Here, pink circles $v$ denotes vertices (or hypernodes) and dashed circles $e$ denotes hyperedges. They are treated equally in this paper.}}
	\label{fig:representation}
	\vspace{-2mm}
\end{figure}

However, the research on spectral graph theory for hypergraphs is far less been developed \cite{ChitraR19}.  Hypergraph learning was first introduced in \cite{ZhouHS06} as a propagation process on hypergraph structure, however, \cite{AgarwalBB06} indicated that their Laplacian matrix is equivalent to pairwise operation. Since then, researchers explored non-pairwise relationships by developing
nonlinear Laplacian operators \cite{chan2018spectral,li2017inhomogeneous}, utilizing random walks \cite{ChitraR19,bellaachia2013random} and learning the optimal weights \cite{li2017inhomogeneous,li2018submodular} of hyperedges. Essentially, most of these algorithms
focus on vertices, viewing hyperedges as connectors, and they
explicitly break the bipartite property of hypergraphs (shown in Figure~\ref{fig:representation}).

Two types of deep learning models have been designed on hypergraphs. \cite{feng2019hypergraph} develops Chebyshev formula for hypergraph Laplacians and proposed HGNN. Using a similar hypergraph Laplacian, \cite{yadati2018hypergcn} proposes HyperGCN while \cite{hyperOperation} generalizes \cite{KipfW17, VelickovicCCRLB18} and defines two neural hypergraph operators. However, this first line of works all construct a simple weighted graph and lose high-order information. A recent work HyperSage \cite{arya2020hypersage} generalizes the message passing neural networks (MPNN) \cite{gilmer2017neural} and uses two-stage message passing functions \cite{dong2020hnhn} on hypergraphs. Based on the two-stage procedure, UniGNN \cite{huang2021unignn} generalizes GCN \cite{KipfW17}, GAT \cite{VelickovicCCRLB18}, GIN \cite{xu2018powerful} models to hypergraphs and AllSet \cite{chien2021you} further unifies a whole class of two-stage models with multiset functions. This type of models are empirically more powerful than the first type, however, they essentially treat the hypergraph as a bipartite graph and build two message passing functions on the heterogeneous structure.

Different from previous works, we propose a novel hypergraph learning model from a new perspective. We propose \emph{line expansion} (\method) for hypergraphs, which is a powerful bijective mapping from a hypergraph structure to a homogeneous graph structure. Relying on our \method mapping, all existing graph representation learning methods \cite{henaff2015deep, KipfW17, DefferrardBV16, VelickovicCCRLB18} can seamlessly and effortlessly work on hypergraphs.

Specifically, from the complex and heterogeneous hypergraph, our \method can induce a simple graph structure (examples in Figure~\ref{fig:expansion}), where the ``node'' is a vertex-hyperedge pair, and ``edge'' between two ``node''s are constructed if two ``node''s share the same vertex or hyperedge (w.l.o.g., we use concept ``node'' in simple graphs and concept ``vertex'' in hypergraphs).  It is interesting that the new induced structure is isomorphic to the line graph of the star expansion of the original hypergraph, which is (i) homogeneous (i.e., a graph where nodes have the same semantics) and (ii) symmetrical with respect to
the vertices and hyperedges. We further prove that \method is also (iii) bijective, which means all high-order information is preserved during the transformation, i.e., the hypergraph can be recovered uniquely from the induced line expansion graph. 

Therefore, to conduct hypergraph representation learning, we {\bf first} transform the hypergraph to the induced simple graph. {\bf Then}, features from hypergraph vertices are projected to node features in the induced graph. {\bf Next}, we apply graph learning algorithms (i.e., graph convolutional network \cite{KipfW17}) to obtain node representations. {\bf Finally}, the node representations from the induced graph is aggregated and back-projected to the original hypergraph vertices for classification. The \method transformation is differentiable and the overall learning process is end-to-end.

The proposed line expansion of hypergraphs is novel and informative. In the traditional formulations, the hyperedges are usually transformed into cliques of edges (e.g., clique/star expansions \cite{AgarwalBB06}) or hypergraph cuts \cite{ZhouHS06}, or the learning solely depends on edge connectivity (e.g., hyperedge expansions \cite{pu2012hypergraph}). Differently, \method treats vertices and hyperedges equally, thus preserving the nature of hypergraphs.
Note that, \method is also significantly different from other hypergraph
formulations, such as tensor based representation \cite{ouvrard2017adjacency}, line graphs of hypergraphs \cite{bermond1977line}, 
intersection graphs of hypergraphs \cite{naik2018intersection}, or middle graphs of hypergraphs
\cite{cockayne1978properties}. These formulations either require strong constraints (e.g., uniform hypergraphs) or result in heterogeneous topologies as well as other structures that complicate practical usage. Previous formulations may restrict applicability of graph algorithms due to their special structures.

Further, this paper revisits the formulation of the standard star or clique expansion and simple graph learning algorithms. We conclude that they can be unified as special cases of \method. 
Empirically, this paper demonstrates the effectiveness of \method 
on five real-world hypergraphs. We apply the popular graph convolutional networks (GCNs)~\cite{KipfW17} on \method as our method, and it consistently outperforms several strong hypergraph learning baselines. 

%

The organization of the paper is as follows. In Section~\ref{sec:pre}, we introduce the general notations of hypergraphs and formulate our problem. In Section~\ref{sec:hrl}, we propose \emph{line expansion} of hypergraphs and show some interesting properties. In Section~\ref{sec:hrl2}, we generalize GCNs to hypergraphs by \emph{line expansion}.
We evaluate \emph{line expansion} on three-fold experiments in Section~\ref{sec:exp}. We conclude this paper and provide proofs for our theoretical statements in the end.


\section{Preliminaries} \label{sec:pre}

\subsection{Hypergraph Notations}

Research on graph-structured deep learning \cite{KipfW17,VelickovicCCRLB18} stems mostly from Laplacian matrix and vertex functions of simple graphs. Only recently, learning on hypergraphs starts to attract attentions from the community \cite{feng2019hypergraph,hyperOperation,chien2021you,arya2020hypersage}.


\smallskip\smallskip
\noindent {\bf Hypergraphs.} Let $\calG_H=(\calV, \calE)$ denote a \emph{hypergraph}, with a vertex set $\calV$ and a edge set $\calE \subset 2^{\calV}$. A hyperedge $e\in \calE$ (we also call it ``edge'' interchangeably in this paper) is a subset of $\calV$. Given an arbitrary set $\calS$, let $|\calS|$ 
denote the cardinality of $\calS$. A simple graph is thus a special case of a hypergraph, with $|e|=2$ uniformly, which is also called a 2-regular hypergraph. 
A hyperedge $e$ is said to be \emph{incident} to a vertex $v$ when $v\in e$. One can represent a 
hypergraph by a $|\calV|\times |\calE|$ \emph{incidence matrix} ${\bfH}$ with its entry $h(v, e) = 1$ if $v\in e$ 
and 0 otherwise. For each vertex $v \in \calV$ and hyperdge $e\in \calE$, $d(v)=\sum_{e\in \calE}h(v, e)$ 
and $\delta(e)=\sum_{v\in \calV} h(v, e)$ denote their degree functions, respectively. The vertex-degree matrix ${\bfD}_v$ of a hypergraph 
$\calG_H$ is a $|\calV|\times |\calV|$ matrix with each diagonal entry corresponding to the node degree, and the edge-degree matrix ${\bfD}_e$ is $|\calE|\times |\calE|$, 
also diagonal, which is defined on the hyperedge degree.  

\subsection{Problem Setup} \label{sec:problem}
Following \cite{feng2019hypergraph, hyperOperation}, this paper studies the transductive learning problems on hypergraphs, specifically node classification,  It aims to induce a labeling $f: \calV\rightarrow \{1, 2, \dots, C\}$ from the labeled data as well as the geometric structure of the graph and then assigns a class label to unlabeled vertices by transductive inference. 

Specifically, given a hypergraph $\calG_H=(\calV, \calE)$ with the labeled vertex set $\calT \subset \calV$ and the labels, we minimize the empirical risk,
\begin{equation}
f^{*}=\underset{f(\cdot \mid\theta)}{\mathrm{arg\, min}}\frac{1}{|\calT|}\sum_{v_t \in \calT} \mathcal{L}(f(v_t\mid \theta),  L(v_t)),
\end{equation}
\noindent
where $L(v_t)$ is the ground truth label for node $v_t$ and cross-entropy error \cite{KipfW17} is commonly applied in $\mathcal{L}(\cdot)$. 

Intuitively, node similarity implies similar labels on the same graph. Given the symmetric structure of hypergraphs, we posit that \emph{vertex similarity} and \emph{edge similarity} are equally important in learning the labels. 
This work focuses on node classification problems on hypergraphs, but it can be easily extended to other hypergraph related applications, such as hyper-edge representation (e.g., relation mining) by exploiting symmetry.


\section{Hypergraph Line Expansion (\method) }\label{sec:hrl}
Most well-known graph-based algorithms \cite{ng2002spectral,grover2016node2vec} are defined for graphs instead of hypergraphs. Therefore, in real-world applications, hypergraphs are
often transformed into simple graphs \cite{ZhouHS06,AgarwalBB06} that are easier to handle.

\subsection{Traditional Hypergraph Expansions} \label{sec:hyperexpansion}
Two main ways of approximating hypergraphs by simple graphs are the clique expansion \cite{sun2008hypergraph} and the star expansion \cite{zien1999multilevel}. The \emph{clique expansion} algorithm (shown in left side of Fig.~\ref{fig:expansion}) constructs a graph $\calG_c=(\calV, \calE_c)$ from the original hypergraph by replacing each hyperedge with a clique in the resulting graph (i.e., $\calE_c = \{(u, v) \mid u, v \in e, e\in \calE\}$), while the \emph{star expansion} algorithm (shown in right side of Fig.~\ref{fig:expansion}) constructs a new graph $\calG_s=(\calV_s, \calE_s)$ by viewing both vertices and hyperedges as nodes in the resulting graph
$\calV_s = \calV \cup \calE$, where vertices and hyperedges are connected by their incident relations (i.e., $\calE_s= \{(v, e)\mid v \in e, v \in \calV, e\in \calE\}$). Note that, the star expansion induces a heterogeneous graph structure.

Unfortunately, these two approximations cannot retain or well represent the higher-order structure of hypergraphs. Let us consider the co-authorship network, as $\calG_H$ in Figure~\ref{fig:expansion}, where we view authors as nodes (e.g., $v_1$, $v_2$) and papers as hyperedges (e.g., $e_1$). Then we immediately know that author $v_1$ and $v_2$ have jointly written one paper $e_1$, and together with author $v_3$, they have another co-authored paper $e_2$. This hierarchical and multi-way connection is an example of \emph{higher-order} relation. Assume we follow the clique expansion, then we obviously miss the information of author activity rate and whether the same persons jointly writing two or more articles.
Though researchers have remedially
used weighted edges \cite{li2017inhomogeneous,ChitraR19}, the hyper-dependency still collapses or fuses into linearity. Star expansion expresses the whole incidence information, but the remaining heterogeneous structure (i) has no explicit vertex-vertex link and (ii) is too complicated for those well-studied graph algorithms, which are mostly designed for simple graphs. One can summarize \cite{HeinSJR13} that these two expansions are not good enough for many real-world applications.

\subsection{Our Line Expansion} \label{sec:lineexpansion}
Since the commonly used expansions cannot give a satisfactory representation, we seek a new expansion that preserves all the original higher-order relations, while presenting an easy-to-learn graph structure.
Motivated by the special symmetric structure of hypergraphs that \emph{vertices are connected to multiple edges
and edges are conversely connected to multiple vertices},  we treat vertices and edges equally and propose hypergraph \emph{Line Expansion} (\method).

\begin{figure}[t]
	\centering
	\includegraphics[width=0.46\textwidth]{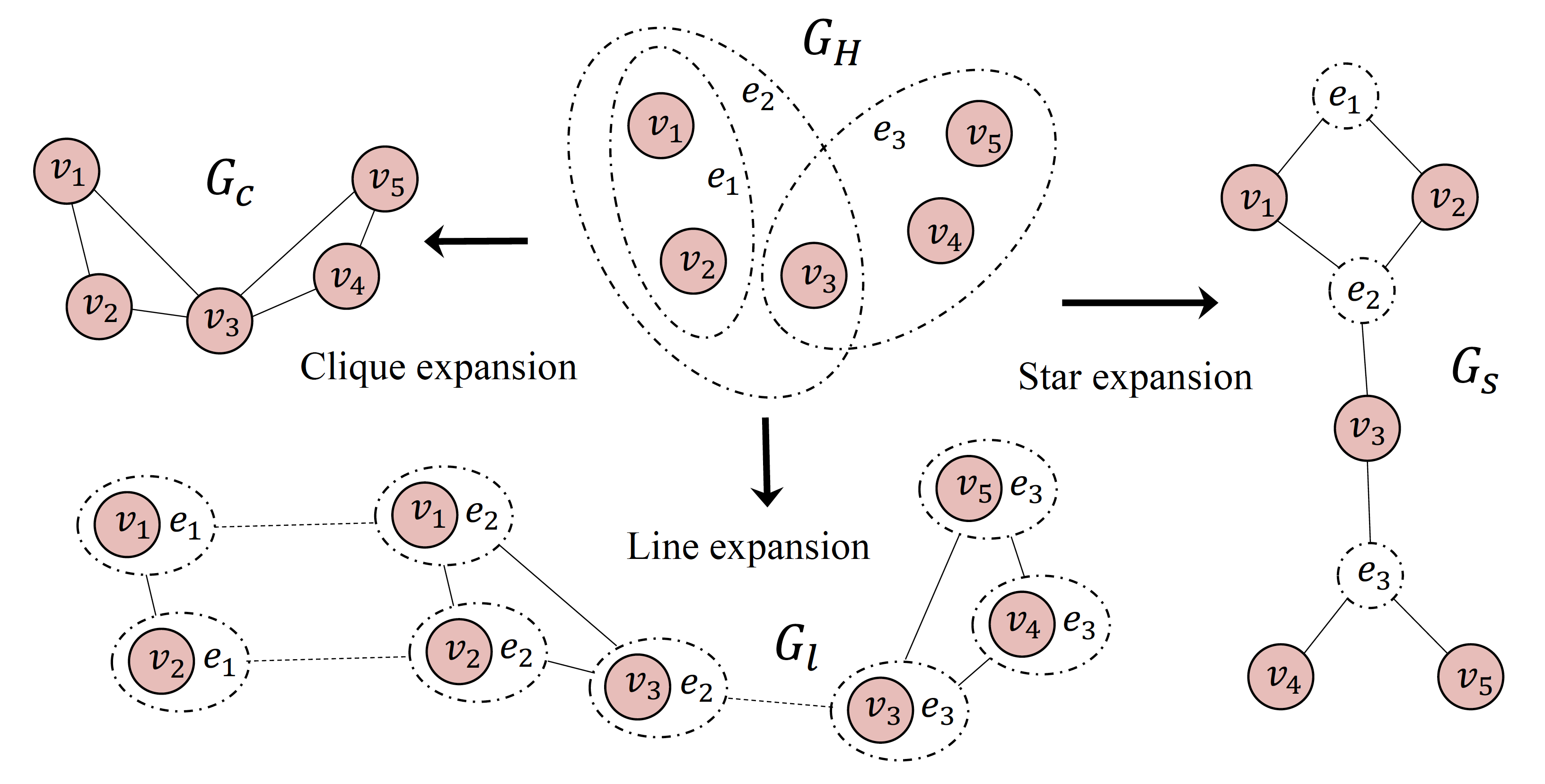}
	\caption{Various Hypergraph Expansions}
	\label{fig:expansion}
\end{figure}

The \emph{Line Expansion} of the hypergraph $\calG_H$ is constructed as follows (shown in Fig.~\ref{fig:expansion}, bottom): (i) each incident vertex-hyperedge pair is considered as a ``line node''; (ii) ``line nodes'' are connected when they share the same vertex or hyperedge. Essentially, the induced structure is a graph, where each vertex or each hyperedge (from the original hypergraph) induces a clique (i.e., fully-connected subgraph). We now formally define the \emph{line expansion}, denoted $\calG_l$.

\subsubsection{Line Expansion.}  Let $\calG_l=(\calV_l,\calE_l)$ denotes the graph induced by the \emph{line expansion} of hypergraph $\calG_H=(\calV,\calE)$.
The node set $\calV_l$ of $\calG_l$ is defined by vertex-hyperedge pair $\{(v, e)\mid v\in e, v\in \calV, e\in \calE\}$  from the original hypergraph. The edge set $\calE_l$  and \emph{adjacency} $\bfA_l\in \{0, 1\}^{|\calV_l|\times |\calV_l|}$  is defined by pairwise relation with $\bfA_l(u_l, v_l)=1$ if either $v=v'$ or $e=e'$ for $u_l=(v, e), v_l=(v', e')\in \calV_l$ . 

The construction of the \emph{line expansion} follows the neighborhood aggregation mechanism. For graph node representation learning, researchers \cite{DefferrardBV16,KipfW17} encode local structure by aggregating information from a
node’s immediate neighborhood. In \emph{line expansion}, we view the incidence of vertex-hyperedge as a whole and generalize the ``neighborhood'' concept by defining that
\emph{two line nodes are neighbors when they contain the same vertex ({vertex similarity}) or the same hyperedge ({edge similarity})}. We argue that
the \emph{line expansion} consequently preserves higher-order associations.

\subsection{Entity Projection}

In this section, we define the projection and back-projection matrices for hypergraph entities (i.e., vertices and hyperedges) between the topological map from $\calG_H=(\calV, \calE)$ to $\calG_l=(\calV_l, \calE_l)$.  In $\calG_l$, each line node $(v,e)$ could be viewed as a vertex with hyperedge context or a hyperedge with vertex context. In a word, the \emph{line expansion} creates information linkage in the higher-order space.  

\smallskip\smallskip
\noindent {\bf Vertex Projection Matrix.} To scatter the information, a vertex $v\in \calV$ from the original hypergraph $\calG_H$ is mapped to a set of line nodes $\{v_l=(v, e):e\in\calE\}\subset \calV_l$ in the induced graph $\calG_l$. We introduce the \emph{vertex projection matrix} $\bfP_{vertex}\in \{0,1\}^{|\calV_l|\times |\calV|}$,
\begin{equation}
\bfP_{vertex}(v_l, v)=\left\{
\begin{aligned}
~1&~  & \mbox{if the vertex part of $v_l$ is $v$},\\
~ 0&~  & otherwise,
\end{aligned}
\right.
\end{equation}
where each entry indicates whether the line node contains the vertex. 

To re-obtain the information of a vertex $v$ in $\calG_H$, we aggregate a set of line nodes in $\calG_l$ who shares the same vertex, for example $v_l=(v,e)$. Since each line node $v_l=(v,e)$ contains the edge context, we consider using the reciprocal of edge size, i.e., $\frac{1}{\delta(e)}$ or $\frac{1}{|e|}$ (check the definition of $\delta(\cdot)$ and $|\cdot|$ in Section~2.1), as the weights for aggregating the line nodes, such that if $\delta(e)$ is smaller (meaning that $v$ is important under the context of $e$), the corresponding line node $(v,e)$ will contribute more to the aggregation.

\smallskip
\noindent {\bf Vertex Back-projection Matrix.}
With this intuition, we fuse the higher-order information by defining the \emph{vertex back-projection matrix} $\bfP_{vertex}'\in \mathbb{R}^{|\calV|\times |\calV_l|}$,
\begin{equation}
\bfP_{vertex}'(v, v_l)=\left\{
\begin{aligned}
~\frac{\frac{1}{\delta(e)}}{\sum_{(v,e')\in\calV_l} \frac{1}{\delta(e')}}&~~~~~  &\mbox{if $v$ is the vertex part of $v_l$}, \\
~ 0&~~~~~  & otherwise.
\end{aligned}
\right.
\end{equation}

Similarly, we could also design \emph{edge projection and back-projection matrices}, $\bfP_{edge}\in \mathbb{R}^{|\calV_l|\times |\calE|}$ and $\bfP'_{edge}\in \mathbb{R}^{|\calE|\times |\calV_l|}$, to exchange information from edges in $\calG_H$ to line nodes in $\calG_l$. 

In fact, the uniqueness of topological inverse mapping from $\calG_l$ to $\calG_H$ is guaranteed by Theorem~\ref{thm:bijective}, where the complete
information of vertex $v\in \calV$ is re-obtained by aggregating all the distributed parts $(v, \cdot)\in \calV_l$ from $\calG_l$.

\begin{thm}\label{thm:bijective} From the line expansion graph, we can uniquely recover the original hypergraph. Formally, the mapping $\phi$ from hypergraph to line expansion (i.e., $\phi: \calG_H \rightarrow \calG_l$) is bijective.
\end{thm}


\subsection{Additional Properties of Line Expansion}\label{sec:renormalized}

In this section, we discuss additional properties of line expansion (\method). First, we present an observation between characteristic matrices from $\calG_H$ and $\calG_l$. Then, we connect our \emph{line expansion} with the line graph from graph theory, based on which, some sound properties could be derived.
 
\begin{observation}\label{thm:observation}
	Let $\bfH$ be the incidence matrix of a hypergraph $\calG_H$. $\bfD_v$ and $\bfD_e$ are the vertex and hyperedge degree matrices. Let $\bfP_{vertex}$ and $\bfP_{edge}$ be the vertex and edge projection matrix, respectively. $\bfA_l$ is the adjacency matrix of line expansion $\calG_l$. Let
	$\bfH_r = \begin{bmatrix} 
	\bfP_{vertex}, \bfP_{edge} \\ 
	\end{bmatrix}\in \{0, 1\}^{ |\calV_l|\times(|\calV|+|\calE|)}$, it satisfies the following equations,
	\begin{align}
	\bfH_r^\top \bfH_r&= \begin{bmatrix} 
	\bfD_v&\bfH\\ 
	\bfH^\top&\bfD_e\\ 
	\end{bmatrix} \label{eq:int1},\\
	\bfH_r \bfH_r^\top &= 2\bfI+ \bfA_l. \label{eq:int2}
	\end{align}
\end{observation}
In Observation~\ref{thm:observation}, the left hand of both Eqn.~\eqref{eq:int1} and Eqn.~\eqref{eq:int2} are the projection matrices, and the right hand of these two equations are information respectively from the hypergraph and its \emph{line expansion}. Essentially, these equations  quantify the transition from $\calG_H$ to $\calG_l$. For Eqn.~\eqref{eq:int2}, we are interested in the product of $\bfH_r \bfH_r ^\top$, leading to two orders of self-loop, which would be useful in the analytical aspects of \emph{line expansion} (in Section~\ref{sec:unification_analysis}).

\begin{thm}\label{thm:linegraph}
	For a hypergraph, its line expansion $\calG_l$ is isomorphic to the line graph of its star expansion $L(\calG_s)$, where $L(\cdot)$ is a line graph notation from graph theory.
\end{thm}

Theorem~\ref{thm:linegraph} is the foundation of Theorem~\ref{thm:bijective}, which provides a theoretical interpretation and enriches our expansion with sound graph theoretical properties (readers could refer to line graph theory \cite{chung1997spectral}). That is why we name our formulation ``line expansion''. Note that the \emph{line expansion} is significantly different from the ``line graph of hypergraph'' discussed in \cite{bermond1977line,bandyopadhyay2020line}. Instead, our line expansion is the line graph of the star expansion. Thus, the proof of Theorem~\ref{thm:linegraph} is naturally established by the construction of line expansion.

Based on Theorem~\ref{thm:linegraph}, we know that $\calG_l$ is homogeneous and has the same connectivity with $\calG_H$. The number of new edges in $\calG_l$ could be calculated as $|\calE_l| = \frac{\sum_{v}d(v)(d(v)-1)}{2}+\frac{\sum_{e}\delta(e)(\delta(e)-1)}{2}$ and line nodes as $|\calV_l|=\frac{\sum_{v}d(v)+\sum_{e}\delta(e)}{2}$. In the worse case, for a fully-connected $k$-regular hypergraph ($k\ll |\calV|$), $|\calV_l|=\Theta(k|\calE|)$ and $|\calE_l|=\Theta(\frac{k^2}{2}|\calE|^2)$, where $\Theta$ is the big Theta notation for the tightest bound. However, many real hypergraphs are indeed sparse (e.g., degrees of vertices and hyperedges follow long-tailed distribution \cite{liu2021tail}, most of them have degree one, $|\calV|\ll |\calE|$ or $|\calE|\ll|\calV|$), so that the scale could usually reduce to $|\calV_l|=\Theta(\frac{|\calV|+|\calE|}{2})$ and $|\calE_l|=O(|\calV||\calE|)$. 


\section{Hypergraph Representation Learning}\label{sec:hrl2}

Transductive learning on graphs is successful due to the fast localization and neighbor aggregation \cite{DefferrardBV16, KipfW17, VelickovicCCRLB18}. It is easy to define the information propagation pattern upon simple structures. For real-world cases, relationships among objects are usually more complex than pairwise. 
Therefore, to apply these algorithms, we need a succinct  informative structure of the higher order relations.

Shown in Section~\ref{sec:hrl}, the bijective map from $\calG_H=(\calV,\calE)$ to $\calG_l=(\calV_l,\calE_l)$ equipped with four entity projectors ($\bfP_{vertex},~\bfP_{vertex}',$ $\bfP_{edge},~\bfP_{edge}'$)
fills the conceptual gap between hypergraphs and graphs. With this powerful tool, it is possible to transfer the hypergraph learning problems into graph structures and address them by using well-studied  graph representation  algorithms. Note that, this work focuses on the generic hypergraphs without edge weights.

\subsection{Hypergraph Learning with Line Expansion} \label{sec:learning}

In this section, we generalize \emph{graph convolution networks} (GCNs) \cite{KipfW17} to hypergraphs and introduce a new learning algorithm defined on \emph{line expansion} for hypergraph representation. Note that, on our proposed structure, other graph representation algorithms could be migrated similarly \cite{perozzi2014deepwalk,tang2015line,HamiltonYL17,VelickovicCCRLB18}.

\subsubsection{Overall Pipeline.} To address the transductive node classification problems on hypergraphs, we organize the pipeline of our proposed model as the following three steps.
\begin{itemize}
    \item STEP1: vertices of the hypergraph is mapped
to line nodes in the induced graph. Specifically, we use the proposed \emph{vertex projection matrix} $\bfP_{vertex}$ to conduct feature mapping. 
\item STEP2: 
we apply deep graph learning algorithms (e.g., GCNs) to learn the representation for each line node. 
\item STEP3: the learned representation is fused by the \emph{vertex back-projection matrix} $\bfP_{vertex}'$ in an inverse edge degree manner. The vertex labels are predicted on the fused representation.
\end{itemize}

\subsection{Convolution on Line Expansion}\label{sec:convolution}

\smallskip
\noindent {\bf STEP 1: Feature Projection.} Given the initial feature matrix $\bfX\in \mathbb{R}^{|\calV|\times d_{i}}$ ($d_i$ is input dimension) from $\calG_H=(\calV,\calE)$,
we transform it into the features in $\calG_l=(\calV_l,\calE_l)$ by vertex projector $\bfP_{vertex}$,
\begin{equation}
\bfH^{(0)}=\bfP_{vertex}\bfX\in \mathbb{R}^{|\calV_l|\times d_i}.
\end{equation}
$\bfH^{(0)}$ is the initial node feature of the induced graph. This projection essentially scatters features from vertex of $\calG_H$ to feature vectors of line nodes in $\calG_l$. In \emph{line expansion} (\method), a line node could be adjacent to another line nodes that contain the same vertex (\emph{vertex similarity}) or the same hyperedge (\emph{edge similarity}).

\smallskip \smallskip
\noindent {\bf STEP 2: Convolution Layer.} We then apply neighborhood feature aggregation by graph convolution. By incorporating information from both vertex-similar neighbors and hyperedge-similar neighbors, the graph convolution is defined as ($k=0, 1,\dots, K$),
\begin{equation}\label{eqn:conv}
h_{(v,e)}^{(k+1)} = \sigma\left(\sum_{e'} w_e h_{(v,e')}^{(k)}\Theta^{(k)}+\sum_{v'} w_v h_{(v',e)}^{(k)}\Theta^{(k)}\right),
\end{equation}
where $h_{(v,e)}^{(k)}$ denotes the feature representation of line node $(v,e)$ in the $k$-th layer, $\sigma(\cdot)$ is a non-linear activation function like \emph{ReLU} \cite{KipfW17} or \emph{LeakyReLU} \cite{VelickovicCCRLB18}. $\Theta^{(k)}$ is the transformation parameters for layer $k$. Two hyper-parameters $w_v, w_e$ are employed to balance \emph{vertex similarity} and \emph{edge similarity}. Specifically, in Eqn.~\eqref{eqn:conv}, the first term (i.e., $\sum_{e'} w_e h_{(v,e')}^{(k)}$) convolves information from neighbors who share the same hyperedges,
whereas the second term (i.e., $\sum_{v'} w_v h_{(v',e)}^{(k)}$) convolves information from neighbors who share the same vertices. 

Eqn.~\eqref{eqn:conv} can be written in matrix version by using the parameterized \emph{adjacency matrix}(In experiment, we use the $w_v=w_e=1$),
\begin{equation}
\bfA_l(u_l, v_l) = \left\{
\begin{aligned}
~w_e&  ~~~~& u_l=(v,e),~v_l=(v',e'),~v=v',\\
~ w_v&  ~~~~& u_l=(v,e),~v_l=(v',e'),~e=e', \\
~0& & ~~~~otherwise,
\end{aligned}
\right.
\end{equation}
and adopt the \emph{renormalized trick} \cite{KipfW17} with the adjustment two-orders of self-loop: $2\bfI +{\bfD_l}^{-\frac12}\bfA_l{\bfD_l}^{-\frac12}\rightarrow \tilde{\bfD_l}^{-\frac12}\tilde{\bfA_l}\tilde{\bfD_l}^{-\frac12}$ (here, $\tilde{\bfA_l} = 2I + \bfA_l$ and $\tilde{{\bfD_l}}_{(ii)}=\sum_j\tilde{{\bfA_l}}_{(ij)}$). Eqn.~\eqref{eqn:conv} can be re-written as,
\begin{equation} \label{eq:hgl}
\bfH^{(k+1)}= \sigma\left(\tilde{\bfD_l}^{-\frac12}\tilde{\bfA_l}\tilde{\bfD_l}^{-\frac12}\bfH^{(k)}\Theta^{(k)}\right), ~~~k=0, 1,\dots, K.
\end{equation}

In real practice, we do not bother to calculate the adjacency $\bfA_l$ directly. An efficient trick is to use Eqn.~\eqref{eq:int2}.

\smallskip
\noindent {\bf STEP 3: Representation Back-projection.} After $K$ layers, $\bfH^{(K)}$ is the final node representation on the induced graph, from which we could derive fused representation for vertices in $\calG_H$. Specifically, we  use the back-projector $\bfP_{vertex}'$,
\begin{equation}\label{eqn:result}
\bfY = \bfP_{vertex}'\bfH^{(K)} \in \mathbb{R}^{|\calV|\times d_o},
\end{equation}
where $d_o$ is
the dimension of output representation.
Note that, in this work, we focus on the 
node classification task. However, due to the symmetry of vertex and edge, this work also sheds some light on the applications of learning hyper-edges (e.g., relation mining) by using $\bfP_{edge}$, $\bfP'_{edge}$. We leave it to future work.

In sum, the complexity of $1$-layer convolution
is of $O(|\calE_l|d_id_o)$, since the convolution operation could be efficiently implemented
as the product of a sparse matrix with a dense matrix.

\subsection{Unifying Hypergraph Expansion} \label{sec:message}


As discussed in Section~\ref{sec:hyperexpansion}, for hypergraphs, common practices often collapse the higher order structure into simple graph structures by attaching weights on edges, and then the vertex operators are solely applied onto the remaining topology. Therefore, the interchangeable and complementary nature between nodes and edges are generally missing \cite{abs-1806-00770}.
\begin{thm}\label{thm:unify}
 Line expansion is a generalization of clique expansion and star expansion. The convolution operator on \method is a generalization of simple graph convolution.
\end{thm}

In this work, instead of designing a local vertex-to-vertex operator \cite{feng2019hypergraph,hyperOperation,ZhouHS06, ZhangHTC17}, we treat the vertex-hyperedge relation as a whole. Therefore, the neighborhood convolution on line expansion is equivalent to exchanging information simultaneously across vertices and hyperedges. Our proposed line expansion (\method) is powerful in that it unifies clique and star expansions, as well as simple graph  cases, stated in Theorem~\ref{thm:unify}. We formulate different hypergraph expansions and provide the proof in Section~\ref{sec:unification_analysis}.

\subsection{Acceleration: Neighborhood Sampling}
For practical usage, we further accelerate the proposed model by neighborhood sampling. As is mentioned, the runtime complexity of our model is proportional to the number of connected edges, $|\calE_l|$, of {\em line expansion}. Real-world hypergraphs are usually sparse, however, most of them often have vertices with large degrees, which would probably lead to a large or dense {\em line expansion} substructure.

On simple graphs, \cite{hamilton2017inductive} proposes neighbor sampling for high-degree vertices, which randomly samples a few neighbors to approximate the aggregation of all neighbors through an unbiased estimator. This paper adopts neighborhood sampling techniques to hypergraph regime and mitigates the potential computational problems in real applications. Since our graph convolution involves both {\em vertex similarity} and {\em edge similarity} information, we design two threshold, $\delta_e$ and $\delta_v$, for two neighboring sets, separately.

We use $\mathcal{N}_E(v,e)$ to denote the hyperedge neighbor set of line node $(v,e)$. Essentially, $\mathcal{N}_E(v,e)$ contains line nodes with same hyperedge context $e$. Similarly, we use $\mathcal{N}_V(v,e)$ as the vertex neighbor set, which contains line nodes with $v$ as the vertex part. For a line node with high ``edge degree", i.e., $|\mathcal{N}_E(v,e)|>\delta_e$, we would randomly sample $\delta_e$ 
elements from $\mathcal{N}_E(v,e)$ to approximate the overall hyperedge neighboring information. Specifically, the first term in Eqn.~\eqref{eqn:conv}, i.e., $\sum_{e'} w_e h_{(v,e')}^{(k)}$, is approximated by,
\begin{equation}
    \sum_{e'} w_e h_{(v,e')}^{(k)}\approx \frac{|\mathcal{N}_E(v,e)|}{\delta_e}\sum^{i=\delta_e}_{i=1:~e_i'\sim\mathcal{N}_E(v,e)} w_e h_{(v,e_i')}^{(k)}.
\end{equation}
Similarly, when $|\mathcal{N}_V(v,e)|>\delta_v$, we would sample $\delta_v$ elements from $\mathcal{N}_V(v,e)$ for the convolution,
\begin{equation}
    \sum_{v'} w_v h_{(v',e)}^{(k)}\approx \frac{|\mathcal{N}_V(v,e)|}{\delta_v}\sum^{i=\delta_v}_{i=1:~v_i'\sim\mathcal{N}_V(v,e)} w_v h_{(v_i',e)}^{(k)}.
\end{equation}
In sum, by adopting the neighbor sampling into hypergraphs, we could effectively reduce the receptive field and prevent the computational problem incurred by high-degree vertices. In the experiments, we empirically show that the running time of our model is comparable to state of the art baselines after sampling.

\begin{table*}[!t]\small
    \caption{Statistics of Hypergraph Datasets}
    	\vspace{-3mm}
	\centering\begin{tabular}{c|cccccc}
		\toprule
		\textbf{Dataset}
		& \textbf{Vertices}  & \textbf{Hyperedges}  &\textbf{Features} & \textbf{Class} & \textbf{Label rate} & \textbf{Training / Validation / Test} \\
		\midrule
		20News     & 16,242  & 100   & 100 & 4    &  0.025 & 400 / 7,921 / 7,921     \\
		Mushroom & 8,124 & 112 & 112 & 2     &  0.006   &  50 / 4,062 / 4,062 \\
		Zoo & 101 & 42  & 17 & 7 & 0.650 & 66 / -- / 35\\
		ModelNet40 & 12,311  & 12,321   & 2048 & 40   &   0.800 &  9,849 / 1,231 / 1,231\\
		NTU2012   & 2,012 & 2,012  & 2048 & 67  & 0.800 &  1,608 / 202 / 202 \\  
		\bottomrule
	\end{tabular}
	\label{tb:hypergraphstat1}
\end{table*}

\begin{table*}[t] \small
	\caption{Structural Complexity Comparison of Baselines and Line Expansion (before and after sampling)}
	\vspace{-3mm}
	\centering\begin{tabular}{ c|ccc|ccc|ccc} 
		\toprule
		{\textbf{Dataset}} & \multicolumn{3}{c|}{\textbf{ * of the clique expansion graph}}&\multicolumn{3}{c|}{\textbf{ * of its line expansion (before)}} &\multicolumn{3}{c}{\textbf{* of its line expansion (after)}} \\
		\cmidrule{2-10}
		& \textbf{Node} & \textbf{Exp. edge} & \textbf{Exp. density} & \textbf{Line node} & \textbf{Line edge} &  \textbf{Density} & \textbf{Line node} & \textbf{Line edge} &  \textbf{Density}\\
		\midrule
		20News     & 16,242 & 26,634,200 & 2.0e-1 & 64,363 & 34,426,427  & 1.6e-2 & 64,363 & 240,233  & 1.2e-4\\
		Mushroom  & 8,124 & 6,964,876 & 2.1e-1 & 40,620 & 11,184,292  & 1.2e-2 & 40,620 & 81,532 & 9.9e-5\\
		Zoo  & 101 & 5,050 & 1.0e-0 & 1,717 & 62,868  & 4.3e-2 & 1,717 & 16,171 & 1.1e-2\\
		ModelNet40   & 12,311 & 68,944 & 9.1e-4 & 61,555 & 317,083  & 1.7e-4 & 61,555 & 310,767  & 1.6e-4\\
		NTU2012   & 2,012  & 10,013 & 4.9e-3 &10,060 & 48,561  & 9.6e-4 & 10,060 & 48,561  & 9.6e-4 \\  
		\bottomrule
	\end{tabular}
	\\
	{Exp. edge} is given by the clique expansion, and Exp. density is computed by $2|E| / |V|(|V| - 1)$ \cite{coleman1983estimation}.
	\label{tb:hypergraphstat2}

\end{table*}

\section{Experiments}\label{sec:exp}
We comprehensively evaluated the proposed {\em line expansion} (\method) with the following experiments and released the implementations\footnote{https://github.com/ycq091044/LEGCN}:
\begin{itemize}
    \item {Real-world hypergraph node classification.}
    \item Special case: simple graph node classification.
    \item Ablation study on the choice of $w_e$ and $w_v$.
\end{itemize} We name our proposed hypergraph learning approach as $\method_{GCN}$.

\subsection{Hypergraph Node Classification}

The main experiment is demonstrated on five real-world hypergraphs with four traditional and four SOTA hypergraph learning methods. The metric is classification accuracy. 


\smallskip
\noindent {\bf Hypergraph Datasets.} The first dataset {\em 20Newsgroups} contains 16,242 articles with binary occurrence values of 100 words. Each word is regarded as a hyperedge and the news articles are vertices. The next two datasets are from the UCI Categorical Machine Learning Repository \cite{Dua:2019}: {\em Mushroom, Zoo}. For these two, a hyperedge is created by all data points which have the same value of categorical features. We follow the same setting from \cite{HeinSJR13} for 20Newsgroups, Mushroom, Zoo (which does not have validate set due to the small scale). Other two are from computer vision/graphics area: {\em Princeton CAD ModelNet40} \cite{wu20153d} and {\em National Taiwan University (NTU) 3D dataset} \cite{chen2003visual}. Though semi-supervised learning usually requires a small training set, we copy the same settings from the original paper \cite{feng2019hypergraph} and use 80\% of the data as training and the remaining 20\% is split into validation and test. The construction of hypergraphs also follows \cite{feng2019hypergraph}. Each CAD model is regarded as a vertex. The formation of hyperedges is by applying MVCNN \cite{su2015multi} on CAD models and then for each model, we assume that its 10 nearest models form a hyperedge.  The initial vertex features are given by GVCNN representations \cite{feng2018gvcnn}. Basic statistics of datasets are reported in Table~\ref{tb:hypergraphstat1}. The graph structure of clique expansion, the line expansion before and after neighbor sampling are reported in Table~\ref{tb:hypergraphstat2}.

\smallskip
\noindent {\bf Baselines.} We select the following baselines. 
\begin{itemize}
    \item {\em Logistic Regression (LR)} works as a standard baseline, which only uses independent feature information. 
    \item {\em $\mbox{Clique}_{GCN}$} and {\em $\mbox{Star}_{GCN}$} are developed by applying GCN on the clique or star expansions of the hypergraphs.
    \item {\em H-NCut} \cite{ZhouHS06}, equivalent to {\em iH-NCut} \cite{li2017inhomogeneous} with uniform hyperedge cost, is a generalized spectral method for hypergraphs. This paper considers {\em H-NCut} as another baseline.
    \item {\em LHCN} \cite{bandyopadhyay2020line} considers the concept of line graph of a hypergraph, however, it irreversibly transforms into a weighted graph.
    \item Hyper-Conv \cite{hyperOperation} and 
    {\em HGNN} \cite{feng2019hypergraph} are two recent models, which uses hypergraph  Laplacians to build the convolution operators.
    \item {\em HyperGCN} \cite{yadati2018hypergcn} approximates each hyperedge by a set of pairwise edges connecting the vertices of the hyperedge. 
\end{itemize}

\begin{table*}[t] \small
	\centering
	\caption{Accuracy and Running Time Comparison on Real-world Hypergraphs (\%)}
	\vspace{-2mm}
	\begin{tabular}{ l|lllll} 
		\toprule
		\textbf{Model} & \textbf{20News}  & \textbf{Mushroom} & \textbf{Zoo} & \textbf{ModelNet40} & \textbf{NTU2012} \\
		\midrule
		LR     & 72.9 $\pm$ 0.7    & 81.6 $\pm$ 0.1   & 74.3 $\pm$ 0.0 & 59.0 $\pm$ 2.8 & 37.5  $\pm$ 2.1 \\
		$\mbox{Star}_{GCN}$ & 68.8  $\pm$ 0.4  & 91.8   $\pm$ 0.3   & 95.2 $\pm$ 0.0  & 90.0 $\pm$ 0.0 & 79.1  $\pm$ 0.0 \\
		$\mbox{Clique}_{GCN}$ & 69.0  $\pm$ 0.3  & 90.0   $\pm$ 0.6   & 94.8 $\pm$ 0.3  & 89.7 $\pm$ 0.4 & 78.9  $\pm$ 0.8 \\
		H-NCut \cite{ZhouHS06}   &72.8 $\pm$ 0.5  & 87.7 $\pm$ 0.2  & 87.3  $\pm$ 0.5& 91.4 $\pm$ 1.1 & 74.8  $\pm$ 0.9 \\  
		LHCN \cite{bandyopadhyay2020line}  & 69.1 $\pm$ 0.4 (37.6s) & 90.2 $\pm$ 0.3 (18.4s)& 55.8 $\pm$ 0.1 (1.8s) & 90.2 $\pm$ 0.2 (145.1s)& 79.9 $\pm$ 0.5 (27.4s)\\
		Hyper-Conv \cite{hyperOperation}  & 73.1  $\pm$ 0.7  (72.6s)  & 93.7   $\pm$ 0.6  (10.6s)   & 93.1 $\pm$ 2.3 (0.8s)  & 91.1 $\pm$ 0.8  (63.1s) & 79.4 $\pm$ 1.3   (6.3s) \\  
		HGNN \cite{feng2019hypergraph} & 74.3  $\pm$ 0.2  (74.2s)  & 93.1 $\pm$ 0.5  (16.1s)   & 92.0 $\pm$ 2.8 (0.8s) & 91.7 $\pm$ 0.4  (61.0s) & 80.0 $\pm$ 0.7    (5.6s) \\
		HyperGCN \cite{yadati2018hypergcn} & 73.6 $\pm$ 0.3  (147.8s)  & 92.3  $\pm$ 0.3  (30.23s)  & 93.1 $\pm$ 2.3 (1.1s) & 91.4 $\pm$ 0.9 (86.5s) & 80.4 $\pm$ 0.7  (9.7s) \\
		\midrule
		$\method_{GCN}$ & \textbf{75.6 $\pm$ 0.2}  (38.6s)   & \textbf{95.2 $\pm$ 0.1}  (18.9s)  & \textbf{97.0 $\pm$ 0.0}  (2.8s) & \textbf{94.1 $\pm$ 0.3}  (85.9s) & \textbf{83.2 $\pm$ 0.2} (9.9s) \\
		\bottomrule
	\end{tabular}
	\label{tb:hypergraphResult}
\end{table*}

\smallskip
\noindent {\bf Experimental Setting.} In the experiment, we set $w_v=w_e=1$ for our model (computation of the adjacency matrix is by  Eqn.~\eqref{eq:int2}). All hyperparameters are selected: GCN with $2$ hidden layers and $32$ units, $50$ as training epochs, $\delta_1=\delta_2=30$ as sampling thresholds, Adam as the optimizer, $2e^{-3}$ as learning rate, $5e^{-3}$ as weight decay, $0.5$ as dropout rate, and $1.5e^{-3}$ as the weight for $L_2$ regularizer. Note that hyperparameters might vary for different datasets, and we specify the configurations per dataset in code appendix. All the experiments are conducted 5 times (to calculate mean and standard deviation) with \textit{PyTorch 1.4.0}  and mainly finished in a 18.04 
LTS Linux server with 64GB memory,  32 CPUs and 4 GTX-2080 GPUs. 


\smallskip
\noindent {\bf Result Analysis.} As shown in Table~\ref{tb:hypergraphResult}, overall our model beat SOTA methods on all datasets consistently. Basically, every model works better than LR, which means transductive feature sharing helps
in the prediction. The performances of traditional $\mbox{Clique}_{GCN}$ and $\mbox{Star}_{GCN}$ are not as good as SOTA hypergraph learning baselines. H-Ncut method depends on linear matrix factorization and it also cannot beat graph convolution methods, which are
more robust and effective with non-linearity. The remaining
four are all graph based deep learning methods, and in essence, they approximate the original hypergraph as a weighted graph and then
utilize vertex functions on the flattened graph. 
The result shows that Our $\method_{GCN}$ is more effective in terms of learning representation and could beat them by $2$\% consistently over all datasets.

\smallskip
\noindent {\bf Discussion of  Complexity.}  We also report the running time comparison in Table~\ref{tb:hypergraphResult}, which already include the neighbor sampling time in our model. Basically, our model is also efficient compared to some state-of-the-arts, especially on {20News, ModelNet40 and NTU2012}, which demonstrates that neighbor sampling does make our proposed model less expensive. Let us investigate this in depth. SOTA hypergraph baselines operate on a
flattened hypergraph (identical to
clique expansion) with designed edge weights. We calculate the number of edges and density for them, denoted as Exp. edge and Exp. density. As shown in Table~\ref{tb:hypergraphstat2}, we find that
the scale of \emph{line expansion} before sampling is within $5$ times of the flattened topology, except for Zoo (flattened topology is a complete graph). However, after sampling the neighbors, line expansion structure has been significantly simplified, and we could also observe that for most of the datasets, the density of the \method graph is much smaller ($\sim\frac{1}{1000}$) than the flattened clique expansion graph.



\subsection{Simple Graph Node Classification}
Since simple graphs are a special case of hypergraphs, $2$-regular hypergraph, we apply \emph{line expansion} to simple graphs to empirically verify our Theorem~\ref{thm:unify} and show that applying graph learning algorithm on line expansion can achieve comparable results.

\smallskip
\noindent {\bf Datasets.} \emph{Cora} dataset has 2,708 vertices and 5.2\% of them have class labels. Nodes contain sparse bag-of-words feature vectors and are connected by a list of citation links. Another two datasets, \emph{Citeseer} and \emph{Pubmed}, are constructed similarly \cite{sen2008collective}. We follow the setting from \cite{yang2016revisiting} and show statistics in Table~\ref{tb:graphstat}.

\begin{table}[tbp]\small
	\centering
	\caption{Statistics of Citation Networks}
	\vspace{-2mm}
	{\begin{tabular}{ c|ccccc} 
			\toprule
			\textbf{Dataset}  & \textbf{Nodes}  & \textbf{Edges}  & \textbf{Features} & \textbf{Class} & \textbf{Label rate} \\
			\midrule
			Cora     & 2,708  & 5,429  & 1,433    & 7       & 0.052      \\
			Citeseer & 3,327  & 4,732  & 4,732    & 6       & 0.036      \\
			Pubmed   & 19,717 & 44,338 & 500      & 3       & 0.003   \\  
			\bottomrule
	\end{tabular}}
	\label{tb:graphstat}
\end{table}

\smallskip
\noindent {\bf Common Graph-based Methods.} We consider the popular deep end-to-end learning methods GCN \cite{KipfW17} and well-known graph representation methods SpectralClustering (SC) \cite{ng2002spectral}, Node2Vec \cite{grover2016node2vec}, DeepWalk \cite{perozzi2014deepwalk} and LINE \cite{tang2015line}. We first directly apply these methods on the simple graphs. Then, we apply them on the line expansion of the simple graph, named $\method_{(\cdot)}$, for example, $\method_{Node2Vec}$.
 Note that, GCNs could input both features and the graph structure (i.e., adjacency matrix), whereas
 other methods only use structural information.

\begin{table}[tbp]\small
\caption{Graph Node Classification Accuracy (\%)}
	\centering
	\vspace{-2mm}
	\begin{tabular}{l|lll}
			\toprule
			\textbf{Model}     & \textbf{Cora}            & \textbf{Citeseer}       & \textbf{Pubmed}  \\
			
			\midrule
			SC & 53.3 $\pm$ 0.2 & 50.8 $\pm$ 0.7 & 55.2 $\pm$ 0.4\\
			  Planetoid & 75.0 $\pm$ 0.9 & 64.0 $\pm$ 1.3 & 76.7 $\pm$ 0.6\\
			  ICA & 74.5 $\pm$ 0.6 & 63.4 $\pm$ 0.6 & 72.9 $\pm$ 1.0 \\
			Node2Vec & 66.3   $\pm$ 0.3 & 46.2  $\pm$ 0.7 &  71.6  $\pm$ 0.5 \\
			DeepWalk     & 62.8 $\pm$ 0.6 & 45.7 $\pm$ 1.2    & 63.4 $\pm$ 0.4\\
			LINE         & 27.7 $\pm$  1.1   & 30.8  $\pm$ 0.2      & 53.5  $\pm$ 0.8\\
			GCN          & 82.6 $\pm$ 0.7 & 70.5 $\pm$ 0.3 &   78.2 $\pm$ 0.6  \\
			\midrule
			$\method_{SC}$ & 56.9 $\pm$ 0.2 & 50.7 $\pm$ 0.2 & 71.9 $\pm$ 0.7\\
			 $\method_{Planetoid}$ & 76.6 $\pm$ 0.4 & 66.0 $\pm$ 0.7 & 77.0 $\pm$ 0.2 \\
			 $\method_{ICA}$ & 72.7 $\pm$  0.4 & 68.6 $\pm$  0.5 & 73.3 $\pm$ 0.7\\
			$\method_{Node2Vec}$ & 74.3 $\pm$ 0.4 & 46.2  $\pm$ 0.1  & 74.3	$\pm$ 0.4\\
			$\method_{DeepWalk}$ & 68.3   $\pm$ 0.1      & 50.4   $\pm$ 0.4     & 68.0 $\pm$ 0.8\\
			$\method_{LINE}$     & 51.7  $\pm$ 0.2        & 34.9  $\pm$ 0.5     & 57.5 $\pm$ 0.3 \\
			$\method_{GCN}$      & 82.3 $\pm$ 0.5& 70.4 $\pm$ 0.3 &   78.7 $\pm$ 0.4  \\
			\bottomrule
	\end{tabular}
	\label{tb:graphresult}
	\vspace{-2mm}
\end{table}

\smallskip
\noindent {\bf Result Analysis.} The accuracy results of node classification for three citation networks are shown in Table~\ref{tb:graphresult}. The experiment clearly demonstrates that \method shows comparable results in graph node classification tasks. Specifically for those non-end-to-end methods, they consistently outperform the original algorithm on simple graphs. The reason might be that \method enriches the plain structure by providing a finer-grained structure and makes nodes edge-dependent, which might explain the improvement in structure-based non-end-to-end models. End-to-end GCNs can reach a much higher accuracy compared to other baselines. We observe that $\method_{GCN}$ tie with original GCN on the three datasets.

\subsection{Ablation Study on on $w_e$ and $w_v$}

In this section, we conduct ablation studies on $w_v$ and $w_e$. Since only the fraction $\frac{w_e}{w_v}$ matters, we symmetrically choose $\frac{w_e}{w_v}  = 0, ~0.1, ~0.2, ~0.5, ~1,$ $~2, ~5, ~10, \infty$ and calculate the test accuracy.

\begin{figure}[t]
	\centering
	\includegraphics[width=0.42\textwidth]{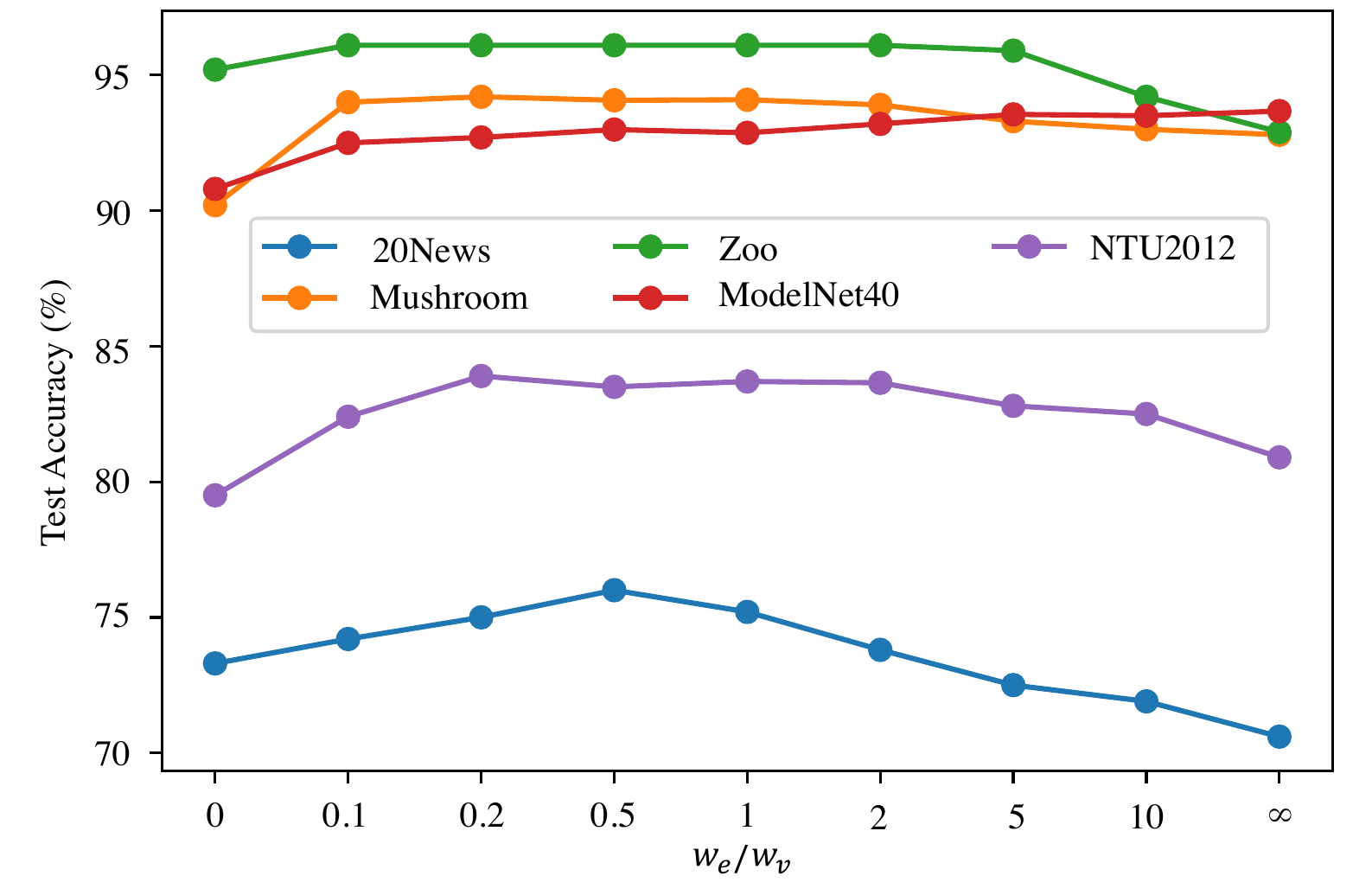}
	\vspace{-2mm}
	\caption{Ablation Study on $\frac{w_e}{w_v}$}
	\label{fig:ablation}
	\vspace{-3mm}
\end{figure}

Figure~\ref{fig:ablation} provides some intuitions on how to select proper $w_v$ and $w_e$ in real hypergraph tasks. We can conclusion that these hypergraphs have different sensitivities and preferences for $w_e$ and $w_v$. However, we do find all the curves follow a first-rise-and-then-down pattern, meaning that it is beneficial to aggregate information from both edge-similar and vertex-similar neighbors. 
Specifically, we find that for hypergraphs with fewer hyperedges, e.g., 20News and Mushroom, the peak appears before $\frac{w_e}{w_v}=1$, and for hypergraphs with sufficient hyperedges, e.g., ModelNet40 and NTU2012, the peak appears after $\frac{w_e}{w_v}=1$. Therefore, one empirical guide for practical usage is to set smaller $w_e$ when there are fewer hyperedges and set larger $w_e$, vice versa. After all, using the binary version (i.e., ${w_e}= {w_v}=1$) seems to be a simple and effective choice.

\section{Conclusion and Future Works}\label{sec:conclusion}

In this paper, we proposed a new hypergraph transformation, \emph{Line Expansion} (\method), which can transform the hypergraph into simple homogeneous graph in an elegant way, without loss of any structure information. With \method, we extend the graph convolution networks (GCNs) to hypergraphs and show that the extended model outperforms strong baselines on five real-world datasets. 

A possible future direction is to exploit the hypergraph symmetry and apply \method for edge learning in complex graphs. Another interesting extension is to extend \emph{line expansion} to directed graphs, where the relations are not reciprocal. In future works, we will further evaluate our model on large hypergraphs, such as DBLP or Yelp, against recent two-stage type hypergraph learning baselines, such as AllSet \cite{chien2021you}, which could be one limitation of the paper.

\section{Proofs} \label{sec:proof}
\vspace{-1mm}
\subsection{Proof of Theorem~\ref{thm:bijective}} \label{proof:thm1}
First, we posit without proof that the hypergraph $\calG_H$ has one-to-one relation with its star expansion $\calG_s$. To prove the bijectivity of mapping $\phi: \calG_H \rightarrow \calG_l$, we can instead prove the bijectivity between $\calG_s$ and $\calG_l$. Our proof will be based on the Whitney graph isomorphism theorem \cite{whitney1992congruent} below.

\begin{figure}[t]
	\centering
	\includegraphics[width=0.3\textwidth]{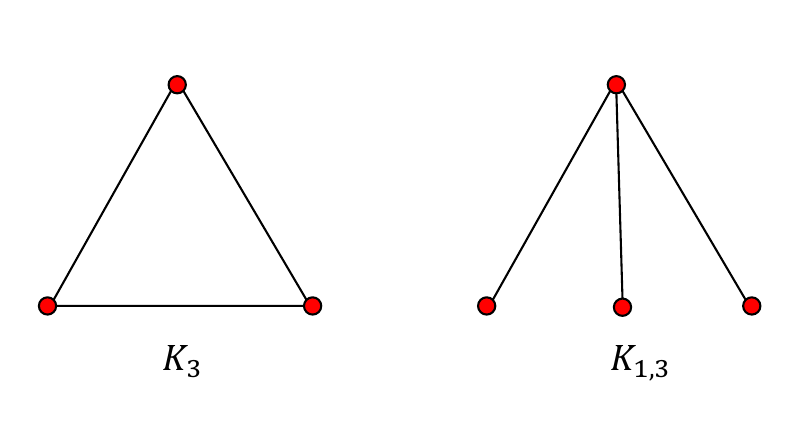}
	\vspace{-4mm}
	\caption{The Exception of Whitney's Theorem}
	\label{fig:appendix2}
\end{figure}
\vspace{-1mm}
\begin{thm} \label{thm:whitney}
(Whitney Graph Isomorphism Theorem.) Two connected graphs are isomorphic if and only if their line graphs are isomorphic, with a single exception: $K_3$, the complete graph on three vertices, and the complete bipartite graph $K_{1,3}$, which are not isomorphic but both have $K_3$ as their line graph.
\end{thm}
\vspace{-2mm}
\begin{definition} \label{def:maximummatching}
(Maximum Independent set.) A maximum independent set is an independent node set (no two of which are adjacent) of largest possible size for a given graph $\calG$.
\end{definition}
\vspace{-2mm}
\begin{proof}For the star expansion of the hypergraph, it could be unconnected when subsets of the vertices are only incident to subsets of the hyperedges. In that case, we could consider the expansion as a union of several disjoint connected components and apply the proof on each component. Below, we mainly discuss the case when
$\calG_s$ is connected.

The proof consists of two
parts. First, we show that for the class of star expansion graphs, Theorem~\ref{thm:whitney} holds without exception. Second,
we show how to recover the star expansion $\calG_s$ (equivalently, the original hypergraph $\calG_H$) from $\calG_l$.

 First, for the exception in Whitney's theorem, it is obvious that $K_3$ (in Figure~\ref{fig:appendix2}) cannot be the star expansion of any hypergraph. Therefore, 
for star expansion graphs (with is also a bipartite representation of the hypergraph), Theorem~\ref{thm:whitney} holds without exception.

\begin{figure}[t]
	\centering
	\includegraphics[width=0.43\textwidth]{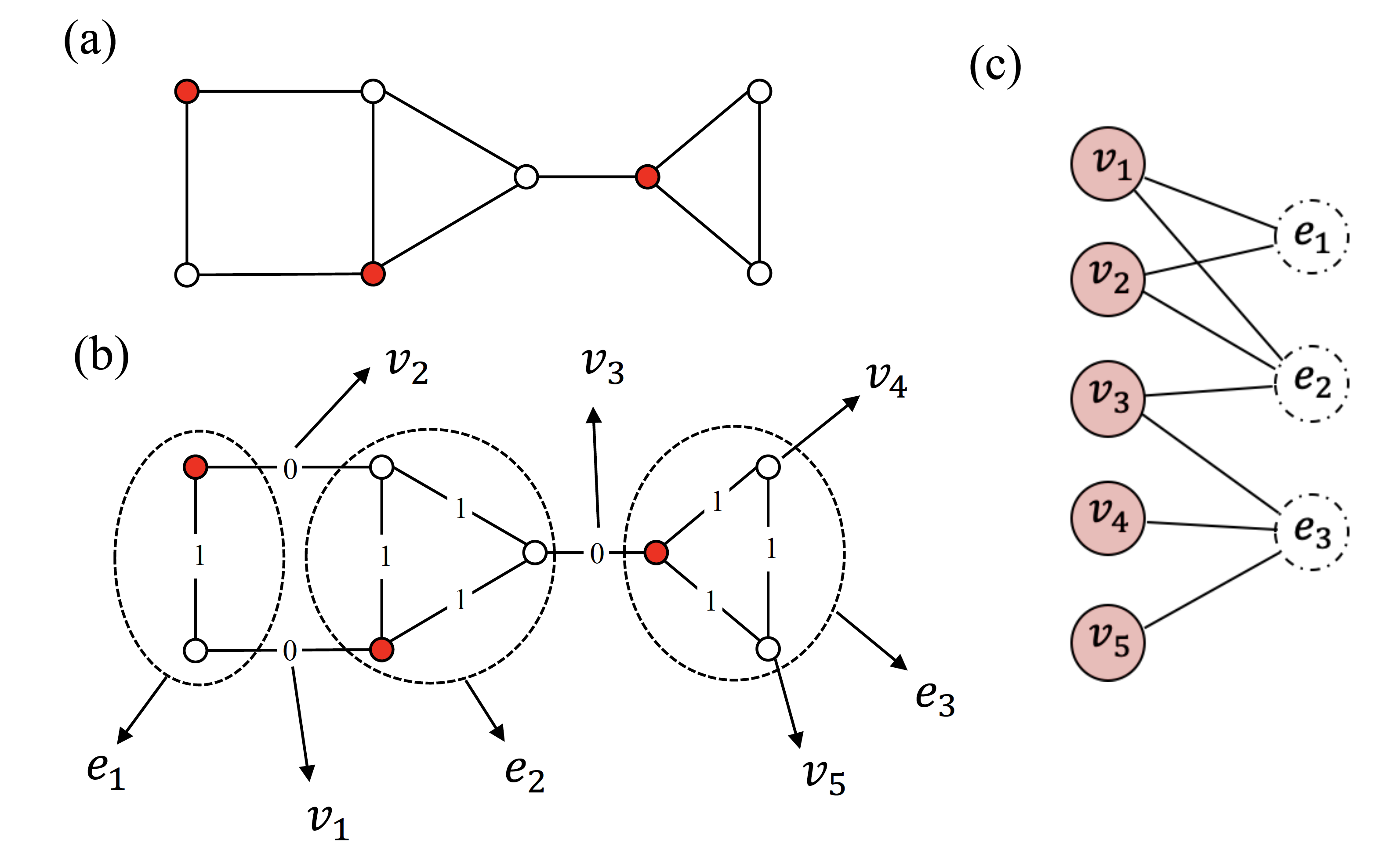}
	\vspace{-4mm}
	\caption{The construction from $G_l$ to $G_s$}
	\vspace{-4mm}
	\label{fig:appendix3}
\end{figure}

Second, given a line graph topology, we know from Theorem~\ref{thm:whitney} immediately that the original bipartite structure is unique.
We now provide a construction from $\calG_l$ to $\calG_s$. Given a
line graph structure, we first find a maximum independent set (in Definition~\ref{def:maximummatching}) and color them in red (shown in Figure~\ref{fig:appendix3} (a)). \cite{paschos2010combinatorial} proves that the maximum independent node could be found in polynomial time. 


Since every vertex and every hyperedge from $\calG_s$ spans a clique in $L(\calG_s)$, let us think
about the line node in the induced graph (which is Fig.~\ref{fig:appendix3}.(b) here), which is potentially a vertex-hyperedge pair. Therefore, each node $(v, e)$ must be connected to exactly two cliques: one spanned by vertex $v$ and
one spanned by hyperedge $e$. Essentially, we try to project these cliques back to original vertex or hyperedges in $G_s$. In fact, for each colored node, we choose one of two cliques connected to it so as to make sure: i) the selected cliques have no intersections (there are two choices. In this case, choose $1$-edge cliques or $0$-edge cliques) and ii) the set of cliques cover all nodes in the topology, shown in Fig.~\ref{fig:appendix3} (b).

For any given line graph topology (of a star expansion graph), we could always find the set of $1$-edge cliques or the set of $0$-edge cliques that satisfies i) and ii), guaranteed by Definition~\ref{def:maximummatching}. Conceptually, due to the bipartite nature, one set will be the cliques
spanned by original hyperedges and another set will be the cliques spanned by original vertices. Either will work for us. Note that the set of $1$-edge cliques also includes two size-1 clique, denoted as $v_4$ and $v_5$ in Fig.~\ref{fig:appendix3} (b).
They seem to only connect to one $1$-edge clique, i.e, $e_3$ clique, however, they are actually size-1 cliques spanned by the original vertices which belongs
to only one hyperedge.

The recovery of the star expansion $\calG_s$ is as follows: First, find a maximum independent set. Second, choose the set of $1$-edge clique and  transform each selected clique as a hyperedge. Then, the vertex set is created two-folded: 
i) a clique with $0$ on its edges is a vertex in $\calG_H$; ii) nodes only connected to one $1$-edge clique are also vertices. By the symmetry of hypergraph, the vertex set and the hyperedge set can also be flipped, but the resulting topology is isomorphic.
\end{proof}

\subsection{Proof of Theorem~3} \label{sec:unification_analysis}
We first formulate the clique, star and our line expansion adjacency matrices and then show the unification evidence.

\noindent {\bf Clique and Star Expansion Adjacency.}
Given a hypergraph $\calG_H=(\calV, \calE)$, consider the clique expansion $\calG_c=(\calV, \calE_c)$. For each pair $(u, v)\in \calE_c$,
\begin{equation}
\bfA_c(u, v) = \frac{w_c(u, v)}{\sqrt{d_c(u)}\sqrt{d_c(v)}}, \label{eq:adjclique}
\end{equation}
where in standard clique expansion, we have,
\begin{align}
w_c(u, v) &= \sum_{}h(u,e)h(v,e), \label{eqn:wc}\\
d_c(u) &=\sum_{}h(u,e)(\delta(e)-1). \label{eqn:dc}
\end{align}

For the same hypergraph $\calG_H=(\calV, \calE)$, star expansion gives $\calG_s=(\calV_s, \calE_s)$. We adopt adjacency formulation from \cite{AgarwalBB06}, formally,
\begin{equation}\label{eqn:star}
\bfA_s(u,v) = \sum_{e\in E}\frac{h(u, e)h(v,e)}{\delta(e)^2\sqrt{\sum_eh(u,e)}\sqrt{\sum_eh(v,e)}}.
\end{equation}

\noindent {\bf Line Expansion Adjacency.}
To analyze the adjacency relation on \emph{line expansion}, we begin by introducing some notations. Let us use
$h^{(k)}_{(v,e)}$ (in short, $h^k_{ve}$) to denote the representation of line node $(v,e)\in V_l$ at the $k$-th layer. The convolution operator on \emph{line expansion}, in main text Eqn.~\eqref{eqn:conv}, can be presented,
\begin{align}\label{eqn:avg1}
h^{k+1}_{ve}
&=\frac{w_e\sum_{e'} h^k_{ve'} + w_v\sum_{v'}  h^k_{v'e}}{w_e(d(v)-1)+w_v(\delta(e)-1)}. 
\end{align}
We augment Eqn.~\eqref{eqn:avg1} by applying 2-order self-loops (mentioned in Section~4.2), and it yields,
\begin{align}\label{eq:operator_on_line}
h^{k+1}_{ve}
&=\frac{w_e\sum_{e'} h^k_{ve'} + w_v\sum_{v'}  h^k_{v'e}}{w_ed(v)+w_v\delta(e)}. 
\end{align}
The above operator is defined on the induced graph, we equivalently convert it into the hypergraph domain by back-projector $\bfP_{vertex}'$. Formally,  assume $x^{k}_u$ as the converted representation for vertex $u$ in hypergraph, Eqn.~\eqref{eq:operator_on_line} can be written as,
\begin{equation}
x^{k+1}_u= \frac{\sum_eh(u,e)\frac{1}{\delta(e)}\frac{w_v\sum_{u'} x^l_{u'} + w_e\sum_{u}  x^l_u}{ w_v\delta(e)+w_ed(u)}}{\sum_eh(u,e)\frac{1}{\delta(e)}}.
\end{equation}
After organizing the equation, we calculate that for each hypergraph vertex pair $(u,v)\in \calV\times \calV$, they are adjacent by,
\begin{equation}
\bfA_l(u,v) =  \frac{\sum_{e}\frac{w_vh(u,e)h(v,e)}{\delta(e)(w_v\delta(e)+w_ed(u))}}{\sum_eh(u,e)\frac{1}{\delta(e)}},
\end{equation}
or by the following form after symmetric re-normalization,
\begin{equation}\label{eqn:linecomplex}
\bfA_l(u,v) =  \frac{\sum_{e}\frac{w_vh(u,e)h(v,e)}{\delta(e)\sqrt{w_v\delta(e)+w_ed(u)}\sqrt{ w_v\delta(e)+w_ed(v)}}}{\sqrt{\sum_eh(u,e)\frac{1}{\delta(e)}}\sqrt{\sum_eh(v,e)\frac{1}{\delta(e)}}}.
\end{equation}

\noindent {\bf Unifying Star and Clique Expansion.} We start by considering the clique expansion graph with weighting function,
\begin{equation}
w_c(u, v) = \sum_{e\in E}\frac{h(u,e)h(v,e)}{(\delta(e)-1)^2}. \label{eqn:newwc}
\end{equation}
Note that this is equivalent to vanish Eqn.~\eqref{eqn:wc} by a factor of $\frac{1}{(\delta(e)-1)^2}$. We plug the value into Eqn.~\eqref{eqn:dc}, 
then adjacency of clique expansion transforms into,
\begin{equation} \label{eqn:cliqueadj}
\bfA_c(u, v) = \frac{\sum_{e}\frac{h(u,e)h(v,e)}{(\delta(e)-1)^2}.}{\sqrt{\sum_{e}h(u,e)\frac{1}{\delta(e)-1}}\sqrt{\sum_{e}h(v,e)\frac{1}{\delta(e)-1}}}.
\end{equation}
Note that when we set $w_e=0$ (no message passing from hyperedge-similar neighbors). The higher-order relation of \emph{line expansion}, in Eqn.~\eqref{eqn:linecomplex} degrades into,
\begin{equation}\label{eqn:linenew}
\bfA_l(u,v) =  \frac{\sum_{e}\frac{h(u,e)h(v,e)}{\delta(e)^2}}{\sqrt{\sum_eh(u,e)\frac{1}{\delta(e)}}\sqrt{\sum_eh(v,e)\frac{1}{\delta(e)}}}.
\end{equation}
Eqn.~\eqref{eqn:linenew} is exactly the adjacency of star expansion in Eqn.~\eqref{eqn:star},
and Eqn.~\eqref{eqn:cliqueadj} (adjacency of clique expansion) is the 1-order self-loop form of the degraded \emph{line expansion}. 

\smallskip
\noindent {\bf Unifying Simple Graph Adjacency.} The convolution operator \cite{KipfW17} on a simple graph can be briefly present,
\begin{equation}
\bfA(u,v)=\frac{\sum_{e}h(u,e)h(v,e)}{\sqrt{d(u)}\sqrt{d(v)}}.\label{eqn:gcn}
\end{equation}
A graph could be regarded as a 2-regular hypergraph, where hyperedge $e$ has exactly two vertices, i.e., $\delta(e)=2$ and each pair of vertices $(u,v)\in \calV\times \calV$ has at most one common edge. Plugging the value into Eqn.~\eqref{eqn:linenew}, and it yields,
\begin{equation}
\bfA_l(u,v) =  \frac{\sum_{e}h(u,e)h(v,e)}{2\sqrt{d(u)}\sqrt{d(v)}}.\label{eqn:linegcn}
\end{equation}
Comparing Eqn.~\eqref{eqn:gcn} and \eqref{eqn:linegcn}, the only difference is a scaling factor $2$, which could be absorbed into filter $\Theta$. 

\subsection{Proof of Observation~1}
First, we have (use $\bfP_v$ to denote $\bfP_{vertex}$ and $\bfP_e$ for $\bfP_{edge}$):
\begin{align}
    \bfH_r^\top \bfH_r = \begin{bmatrix}
     \bfP_v^\top \\
     \bfP_e^\top \\
    \end{bmatrix}\begin{bmatrix}
     \bfP_v  \bfP_e
    \end{bmatrix}
    = \begin{bmatrix}
     \bfP_v^\top  \bfP_v &  \bfP_v^\top  \bfP_e \\
     \bfP_e^\top  \bfP_v &  \bfP_e^\top  \bfP_e \\
    \end{bmatrix}
    = \begin{bmatrix} 
	 \bfD_v& \bfH\\ 
	 \bfH^\top& \bfD_e\\ 
	\end{bmatrix},
\end{align}
where the last equality is easy to verify since i) $ \bfP_v^\top  \bfP_v$ implies the vertex degree matrix, which is $ \bfD_v$. ii) $ \bfP_e^\top  \bfP_e$ implies the hyperedge degree matrix, which is $ \bfD_e$; iii) $ \bfP_v^\top  \bfP_e$ implies the vertex-hyperedge incidence,
which is $ \bfH$.

For Eqn.~\eqref{eq:int2}, each row of $ \bfH_r$ is a $0/1$ vector of size $|\calV|+|\calE|$ with each dimension indicating a vertex or a hyperedge.  Therefore, the vector has exactly two $1$s, which is due to that a line node contains exactly one vertex and one hyperedge. 

For the $(i,j)$-th entry of $ \bfH_r  \bfH_r^\top$, it is calculated by the dot product
of row $i$ (line node $i$) and row $j$ (line node $j$) of $\bfH_r$. If $i=j$, then this entry will be $2$ (dot product of the same $0/1$ vector with two $1$s). If $i\neq j$, the result will be $0$ if line node $i$ and line node $j$ has no common vertex or hyperedge and be $1$ if they share vertex or hyperedge (the corresponding dimension gives $1$ and $0$ for other dimensions, summing to $1$). In sum,
$\bfH_r \bfH_r^\top$ is equal to the adjacency $\bfA_l$ with 2-order self-loops, quantitatively,
\begin{equation}
    \bfH_r \bfH_r^\top = 2\bfI+ \bfA_l.
\end{equation}


\section*{Acknowledgements}

This work was in part supported by DARPA award HR001121C0165, DoD Basic Research Office award HQ00342110002, NSF grants IIS-2107200 and CPS-2038658.

\bibliographystyle{ACM-Reference-Format}
\balance
\bibliography{sample-base}


\end{document}